\def\year{2019}\relax
\documentclass[letterpaper]{article}
\usepackage{aaai19}
\usepackage{times}
\usepackage{helvet}
\usepackage{courier}
\usepackage{url}
\usepackage{graphicx}
\usepackage{amssymb}
\usepackage{subfig}
\usepackage{amsmath}
\usepackage{mathrsfs}
\DeclareMathAlphabet{\mathpzc}{OT1}{pzc}{m}{it}

\begin{document}
%
\title{Entropy Based Independent Learning in Anonymous Multi-Agent Settings}
\author{Tanvi Verma, Pradeep Varakantham and Hoong Chuin Lau\\
	School of Information Systems, Singapore Management University, Singapore \\
	tanviverma.2015@phdis.smu.edu.sg, pradeepv@smu.edu.sg, hclau@smu.edu.sg }
\maketitle
\begin{abstract}
Efficient sequential matching of supply and demand is a problem of interest in many online to offline services. For instance, Uber, Lyft, Grab for matching taxis to customers; Ubereats, Deliveroo, FoodPanda etc for matching restaurants to customers. In these online to offline service problems, individuals who are responsible for supply (e.g., taxi drivers, delivery bikes or delivery van drivers) earn more by being at the "right" place at the "right" time. We are interested in developing approaches that learn to guide individuals to be in the "right" place at the "right" time (to maximize revenue) in the presence of other similar "learning" individuals and only local aggregated observation of other agents states (e.g., only number of other taxis in same zone as current agent).

Existing approaches in Multi-Agent Reinforcement Learning (MARL) are either not scalable (e.g., about 40000 taxis/cars for a city like Singapore) or assumptions of common objective or action coordination or centralized learning are not viable. 
A key characteristic of the domains of interest is that the interactions between individuals are anonymous, i.e., the outcome of an interaction (competing for demand) is dependent only on the number and not on the identity of the agents. We model these problems using the Anonymous MARL (AyMARL) model.  To ensure scalability and individual learning, we focus on improving performance of independent reinforcement learning methods, specifically Deep Q-Networks (DQN) and Advantage Actor Critic (A2C) for AyMARL. The key contribution of this paper is in employing principle of maximum entropy to provide a general framework of independent learning that is both empirically effective (even with only local aggregated information of agent population distribution) and theoretically justified.  

Finally, our approaches provide a significant improvement with respect to joint and individual revenue on a generic simulator for online to offline services and a real world taxi problem over existing approaches.  More importantly, this is achieved while having the least variance in revenues earned by the learning individuals, an indicator of fairness. 
\end{abstract}

\section{Introduction}
Systems that aggregate supply to ensure a better matching of demand and supply have significantly improved performance metrics in taxi services, food delivery, grocery delivery and other such online to offline services. These systems continuously and sequentially match available supply with demand. We refer to these problems as Multi-agent Sequential Matching Problems (MSMPs). For the individuals (taxi drivers, delivery boys, delivery van drivers) that are responsible for supply, learning to be at the "right" place at the "right" time is essential to maximizing their revenue. This learning problem is challenging because of:
\begin{itemize}
\item Uncertainty about demand (which will determine the revenue earned) 
\item To ensure business advantage over competitors, centralized applications (e.g., Uber, Deliveroo, Lyft etc.) limit access for individuals to only aggregated and local information about the locations of other individuals (which will determine whether the demand is assigned to the current individual).
\end{itemize}

Given the presence of multiple selfish learning agents in MSMPs, Multi-Agent Reinforcement Learning (MARL) \cite{tan1993multi,busoniu2008comprehensive} is of relevance. In MARL, multiple agents are learning to maximize their individual rewards. There are two main issues with using MARL approaches in case of  sequential matching problems of interest: (a) The definition of fairness when learning for multiple agents is unclear as the agent actions may or may not be synchronised. Furthermore, agents can have different starting beliefs and demand is dynamic or non-stationary. (b) The scale of matching problems is typically very large (e.g., 40000 taxis in a small city like Singapore) and hence joint learning is computationally infeasible with no guarantee of convergence (due to non-stationarity).     

An orthogonal thread of MARL has focussed on teams of agents that maximize the global reward~\cite{perolat2017multi,nguyen2017policy} or agents cooperate to achieve a common goal \cite{hausknecht2015deep,foerster2016learning} or make assumptions of centralized learning~\cite{nguyen2017policy}. The learning approaches are also limited in scalability with respect to the number of agents or the types of agents. More importantly, the problems of interest in this paper are focused on individual agents that are maximizing their own revenue rather than a team revenue and hence centralized learning is not viable. Therefore, this thread of MARL approaches are not suitable for MSMPs of interest in this paper.

Existing work~\cite{verma2017rltaxi} in MSMPs has employed Reinforcement Learning (RL) for each individual to learn a decision making policy. However, this line of work has assumed that other agents employ stationary and fixed strategies, rather than adaptive strategies. Due to the non-stationarity introduced when multiple agents learn independently, the performance of traditional table based Q-learning approaches is extremely poor. 

The use of deep neural networks has become hugely popular in the RL community \cite{mnih2015human,silver2017mastering}. 
In this paper, we improve the leading independent RL methods, specifically DQN (Deep Q-Networks) and A2C (Advantage Actor Critic) for solving problems of interest. Specifically:
\begin{itemize}
\item Extending on past work on interaction anonymity in multi-agent planning under uncertainty~\cite{varakantham2012decision,varakantham2014decentralized,nguyen2017collective}, we describe AyMARL for representing interaction anonymity in the context of MARL.  
\item To account for limited information about other agents states and actions while exploiting interaction anonymity, we  develop a framework for independent learning (in AyMARL) that employs principle of maximum entropy . 
\item We then use the independent learning framework to extend the leading single agent reinforcement learning approaches namely DQN (Deep Q-Network) and A2C (Advantage Actor-Critic) methods.  
\item To demonstrate the utility of our approaches, we performed extensive experiments on a synthetic data set (that is representative of many problems of interest) and a real taxi data set. We observe that our individual learners based on DQN and A2C are able to learn policies that are fair (minor variances in values of learning individuals) and most importantly, the individual and joint (social welfare) values of the learning agents are significantly higher than the existing benchmarks. 
\end{itemize}

\section{Motivating Problems (MSMPs)}
\label{motivating}

This paper is motivated by Multi-agent Sequential Matching Problems (MSMPs) where there are multiple agents and there is a need for these agents to be matched to customer demand. Aggregation systems (Uber, Lyft, Deliveroo etc.) maximize the overall system wide revenue in MSMPs.  A key characteristic of these domains is that interactions between individual agents are anonymous. While there are many such domains, here we describe three popular MSMPs: 

\noindent \textbf{Taxi Aggregation:}
Companies like Uber, Lyft, Didi, Grab etc. all provide taxi supply aggregation systems. The goal is to ensure wait times for customers is minimal or amount of revenue earned is maximized by matching taxi drivers to customers.  However, these goals of the aggregation companies may not be correlated to the individual driver objective of maximizing their own revenue. The methods provided in this paper will be used to guide individual drivers to "right" locations at "right" times based on their past experiences of customer demand and taxi supply (obtained directly/indirectly from the aggregation company), while also considering that other drivers are learning simultaneously to improve their revenue. Interactions between taxi drivers are anonymous, because the probability of a taxi driver being assigned to a customer is dependent on the number of other taxi drivers being in the same zone (and not on specific taxis being in the zone) and customer demand.

\noindent \textbf{Food or Grocery Delivery:}
Aggregation systems have also become very popular for food delivery (Deliveroo, Ubereats, Foodpanda, DoorDarsh etc.) and grocery delivery (AmazonFresh, Deliv, RedMart etc.) services. They offer access to multiple restaurants/grocery stores to the customers and use services of delivery boys/delivery vans to deliver the food/grocery.  
 Similar to taxi case, there is anonymity in interactions as the probability of a delivery boy/van being assigned a job is dependent on number of other delivery boys/vans being in the same zone and customer demand. 

\noindent \textbf{Supply Aggregation in Logistics:}
More and more on-line buyers now prefer same day delivery services and tradition logistic companies which maximize usage of trucks, drivers and other resources are not suitable for it. Companies like Amazon Flex, Postmates, Hitch etc. connect shippers with travelers/courier personnel to serve same day/on-demand delivery requests. The courier personnel in this system can employ the proposed method to learn to be at "right" place at "right" time by learning from the past experiences. Interactions between couriers are anonymous due to dependence on number of other couriers (and not on specific couriers).

\section{Related Work}
Researchers have proposed various methods to tackle the non-stationarity of the environment due to the presence of multiple agents. \citeauthor{foerster2017stabilising}\shortcite{foerster2017stabilising} provided multi-agent variant of importance sampling and proposed use of fingerprint to track the quality of other agents' policy to tackle the non-stationarity. \citeauthor{lanctot2017unified} \shortcite{lanctot2017unified} used a game-theoretic  approach to learn the best responses to a distribution over set of policies of other agents. 
Learning with Opponent-Learning Awareness (LOLA) was introduced by \citeauthor{foerster2017learning} \shortcite{foerster2017learning} which explicitly accounts for the fact that the opponent is also learning. 
Another thread of work in multi-agent domains focus of centralized training with decentralized execution \cite{lowe2017multi,nguyen2017policy,pmlrv80yang18d}, where actor-critic method is used with a central critic which can observe the joint state and actions of all the agents. These approaches require access to more information such as joint state, joint action, opponents' parameters, a centralized payoff table etc.,  which is not feasible for our problem domain where there are self interested agents. 

A comprehensive survey on MARL dealing with non-stationarity has been provided by \citeauthor{hernandez2017survey} \shortcite{hernandez2017survey} where based on how the algorithms cope up with the non-stationarity they are categorised into five groups: ignore, forget, respond to target opponents, learn opponent models and theory of mind. The first two categories do not represent other agents, while  the last three categories explicitly represent other agents (in increasing order of richness in models used to represent other agents). Our approach falls in the third category, where we respond to a summary (represented using the agent population distribution) of target opponent strategies. 

Mnih {\em et al.}~\shortcite{mnih2016asynchronous} and Haarnoja {\em et al.}~\shortcite{haarnoja2017soft} use policy entropy regularizer to improve performance of policy gradient approaches to Reinforcement Learning. Our approaches are also based on the use of entropy, however, there are multiple significant differences.  Policy entropy is entropy on policy that is relevant for single agent learning. We employ density entropy, which is entropy on joint state configuration in multi-agent settings. Unlike policy entropy which is a regularization term, considering density entropy requires fundamental changes to the network and loss functions (as explained in future sections). Policy entropy is used to encourage exploration, while density entropy helps in improving predictability of joint state configuration and reduces non-stationarity (by moving all learning agents to high entropy joint states) due to other agents' strategies. Finally, since policy entropy is for single agent exploration, it is complementary to density entropy. 

\section{Background: Reinforcement Learning}

In this section, we briefly describe the Reinforcement Learning (RL) problem and leading approaches (of relevance to this paper) for solving large scale RL problems. 

The RL problem for an agent is to maximize the long run reward while operating in an environment represented as a Markov Decision Process (MDP). Formally, an MDP is represented by the tuple $\big \langle {S,A,T,R }\big \rangle$, where ${S}$ is the set of states encountered by the agent, $ {A}$ is the set of actions, ${T}(s,a,s')$ represents the probability of transitioning from state $s$ to state $s'$ on taking action $a$ and ${R}(s,a)$ represents the reward obtained on taking action $a$ in state $s$. The RL problem is to learn a policy that maximizes the long term reward while only obtaining experiences (and not knowing the full reward and transition models) of transitions and reinforcements. An experience is defined as $(s, a, s', r)$. 

We now describe three approaches that are of relevance to the work described in this paper. 

\subsection{Q-Learning}

One of the most popular approaches for RL is Q-learning~\cite{watkins1992q}, where the Q function is represented as a table (and initialised to arbitrary fixed values) with an entry in the table for each state and action combination. Q-values are updated based on each experience given by $(s,a,s',r)$:
{\small  \begin{align}
{Q}(s,a) \leftarrow  {Q}(s,a) + \alpha [r + \gamma \cdot \max_{a'} Q (s',a') - Q(s,a)] \label{eqn:1}
\end{align} }
Where $\alpha$ is the learning rate and $\gamma$ is the discount factor. To ensure a good explore-exploit tradeoff, an $\epsilon$-greedy policy is typically employed. Q-learning is guaranteed to converge to the optimal solution for stationary domains~\cite{watkins1992q}. 

\subsection{Deep Q-Networks (DQN)}
\label{bg:dqn}

Instead of a tabular representation for Q function employed by Q-Learning, the DQN approach \cite{mnih2015human} employs a deep network to represent the Q function. Unlike with the tabular representation that is exhaustive (stores Q-value for every state, action pair), a deep network predicts Q-values based on similarities in state and action features. This deep network for Q-values is parameterized by a set of parameters, $\theta$ and the goal is to learn values for $\theta$ so that a good Q-function is learnt. 

We learn $\theta$ using an iterative approach that employs gradient descent on the loss function. Specifically, the loss function at each iteration is defined as follows:
{ \begin{align*}
\mathcal{L}_{\theta} = \mathbb{E}_{(e \sim U(\mathcal{J}))} [(y^{DQN} - Q(s,a;\theta))^2]
\end{align*}}
where $y^{DQN} = r + \gamma \text{max}_{a'} Q(s', a'; \theta^-)$ is the target value computed by a target network parameterized by previous set of parameters, $\theta^-$. Parameters $\theta^-$ are frozen for some time while updating the current parameters $\theta$. To ensure independence across experiences (a key property required by Deep Learning methods to ensure effective learning), this approach maintains a replay memory ${\mathcal J}$ and then experiences $e \sim U(\mathcal{J})$ are drawn uniformly from the replay memory. 
Like with traditional Q-Learning, DQN also typically employs an $\epsilon$-greedy policy. 

\subsection{Advantage Actor Critic (A2C)}
\label{bg:a2c}

DQN learns the Q-function and then computes policy from the learnt Q-function. A2C is a policy gradient method that directly learns the policy while ensuring that the expected value is maximized. To achieve this, A2C employs two deep networks, a policy network to learn policy, $\pi(a|s;\theta_{p})$ parameterized by $\theta_{p}$ and a value network to learn value function of the computed policy, $V^{\pi}(s; \theta_v)$ parameterized by $\theta_{v}$. 

The policy network parameters are updated based on the policy loss, which in turn is based on the advantage ($A(s,a;\theta_{v})$) associated with the value function. Formally, the policy loss is given by:
{ 
\begin{align}
{\mathcal L}_{\theta_{p}} &= \nabla_{\theta_{p}} \text{log} \pi(a|s;\theta_{p}) A(s,a;\theta_{v}) \quad \text{where,} \nonumber \\
&A(s,a;\theta_{v}) = R - V(s;\theta_{v}) \label{adv}
\end{align}}
Here $R$ is the $k$-step discounted reward following the current policy and using a discount factor $\gamma$.

\section{Anonymous  Multi-Agent Reinforcement Learning (AyMARL)}
\begin{figure}
    \centering
   \centerline{\includegraphics[scale=0.3]{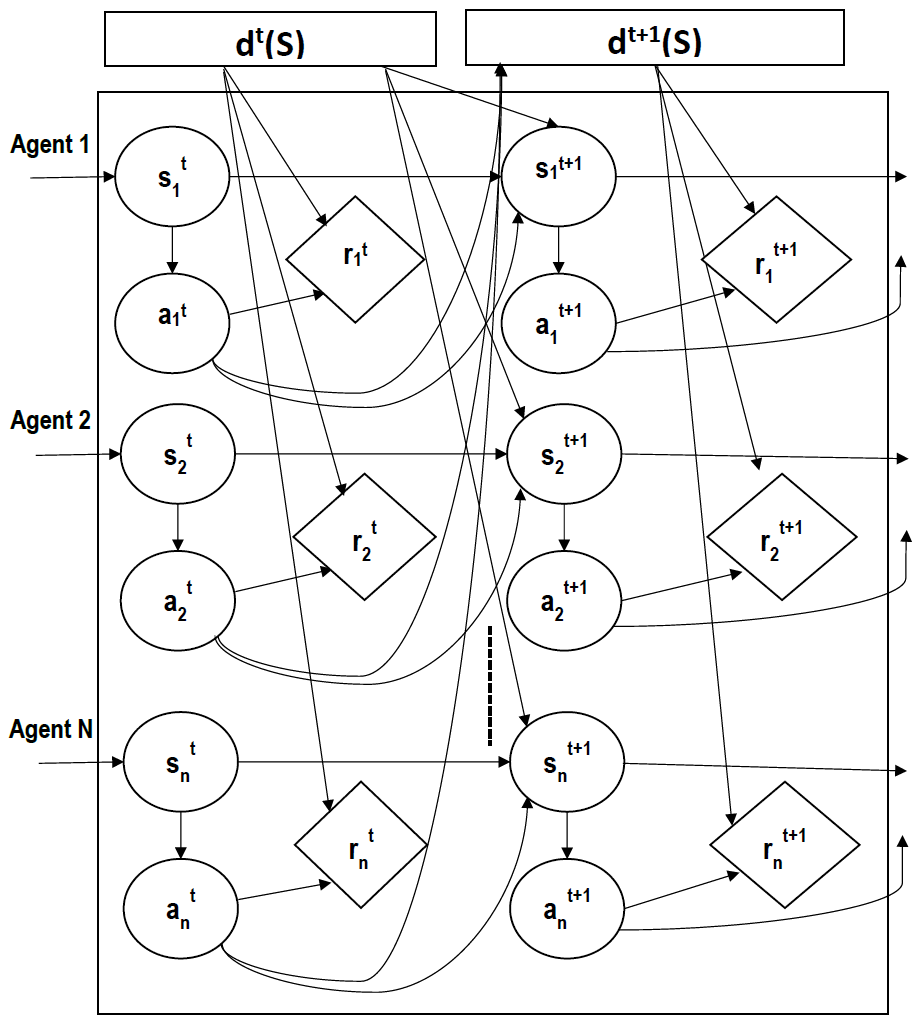}}
   \caption{Agent population distribution $\textbf{d}^t(\textbf{s})$ and action $a_i^t$ affects reward $r_i^t$ of agent $i$. Distribution $\textbf{d}^{t+1} (\textbf{s})$ is determined by the joint action at time step $t$.}
\label{dbn}
\end{figure}
The generalization of MDP to the multi-agent case with special interested agents is the stochastic game model~\cite{shapley1953stochastic}. The underlying model in AyMARL is a specialization of the stochastic game model that considers interaction anonymity and is represented using the Dynamic Bayes Network (DBN) in Figure~\ref{dbn}. Formally, it is represented using the tuple: 
$$\left<\mathcal{N}, \mathcal{S}, \{\mathcal{A}_{i} \}_{i \in \mathcal{N}}, \mathcal{T}, \{\mathcal{R}_i\}_{i \in \mathcal{N}} \right>$$
$\mathcal{N}$ is the number of agents. $\mathcal{S}$ is the set of states, which is factored over individual agent states in MSMPs. For instance in the taxi problem of interest, an environment state $\textbf{s}$ is the vector of locations (zone) of all agents, i.e., $\textbf{s} (\in \mathcal{S}) =  (z_1, \ldots, z_i, \ldots, z_{\mathcal N})$.
${\mathcal{A}}_i$ is the action set of each individual agent. In case of the taxi problem, an action $a_i (\in \mathcal{A}_i)$ represents the zone to move to if there is no customer on board.  

\noindent $\mathcal{T}$ is the transitional probability of environment states given joint actions. 
{\small \begin{align*}
&p(\textbf{s}'|\textbf{s}, \textbf{a}) = \mathcal{T}(\textbf{s},\textbf{a},\textbf{s}') = \mathcal{T}(\textbf{s}, (a_1,\ldots, a_{\mathcal{N}}), \textbf{s}')
\end{align*}}
As shown in the DBN of Figure~\ref{dbn}, individual agent transitions are not dependent on other agents' states and actions directly but through agent population distribution, $\textbf{d}$. That is to say, given $\textbf{d}$, individual agent transitions are independent of each other in AyMARL. 

{ \small \begin{align}
p(\textbf{s}'|\textbf{s}, \textbf{a})= \prod_{i} \mathcal{T}_i(z_i, a_i, z'_i, d_{z_i}(\textbf{s}))  
 \label{eqn:prob}
\end{align}}
where $d_{z_i} (\textbf{s}) = \#_{z_i}(\textbf{s})$, is the number of agents in zone $z_i$ in joint state $\textbf{s}$ and $\textbf{d}(\textbf{s}) = <d_{z_1} (\textbf{s}),..., d_{z_i} (\textbf{s}),...d_{z_\mathcal{N}} (\textbf{s})>$.  Equation \ref{eqn:prob} can be rewritten in terms of $p(\textbf{d}'(\textbf{s}'_{-i}) | \textbf{d}(\textbf{s}), \textbf{a}_{-i})$, which is the probability of other agents (except agent $i$) having an agent population distribution $\textbf{d}'(\textbf{s}'_{-i})$ given current distribution $\textbf{d}(\textbf{s})$ and joint action of the other agents $\textbf{a}_{-i}$
{\small \begin{align}
 p(\textbf{s}'|\textbf{s}, \textbf{a})=\mathcal{T}_{i}(z_i,a_i,z'_i, d_{z_i}(\textbf{s})) \cdot p(\textbf{d}'(\textbf{s}'_{-i}) | \textbf{d}(\textbf{s}), \textbf{a}_{-i}) \label{eqn:tran}
\end{align}}
 
\noindent $\mathcal{R}_i(\textbf{s}, a_i)$ is the reward obtained by agent $i$ if the joint state is $\textbf{s}$ and action taken by $i$ is $a_i$. Like with transitions, as shown in DBN, agent rewards are independent given agent population distribution, $\textbf{d}$. In AyMARL,  it is given by $\mathcal{R}_i(z_i, a_i, d_{z_i}(\textbf{s}))$. 

\textbf{Extensions to AyMARL: }  For purposes of explainability, we will employ the  model described above in the rest of the paper. However, we wish to highlight that multiple extensions are feasible for AyMARL and can be readily addressed with minor modifications to learning approaches.  First extension is the dependence of transition and reward on $\textbf{d}(\textbf{s})$ (or some subset of $\textbf{d}$ given by $d_{\{z_j,z_k,\ldots,\}}(\textbf{s})$) and not just on $d_{z_i}(\textbf{s})$. A second extension is considering dependence of transition and reward on agent state, action distributions and not just on agent population distributions.  That is to say, reward and transitions are dependent on $d_{z_i, a_i} (\textbf{s}, \textbf{a})$ or more generally $\textbf{d}(\textbf{s},\textbf{a})$.

\section{Entropy based Independent Learning in AyMARL}

Due to the independent and selfish nature of agents (maximizing their own revenue) coupled with the lack of access to specific state and action information of other agents, we focus on independent learning approaches for AyMARL problems. In this section, we provide mathematical intuition and a general framework for independent learning based on the use of principle of maximum entropy in settings of interest. 
 
 Q-function expression for a given agent $i$ in stochastic games \cite{hu2003nash} is given by: 
{\small \begin{align}
\mathcal{Q}_i(\textbf{s}, \textbf{a}) &= {\mathcal R}_i(\textbf{s}, a_i) + \gamma \cdot \sum_{\textbf{s}'} p(\textbf{s}'| \textbf{s}, \textbf{a}) \cdot \max_{\textbf{a}'} \mathcal{Q}_i(\textbf{s}', \textbf{a}') \label{QSG}
\end{align}}

In MSMPs of interest, individual agents (e.g., taxis, truck drivers or delivery boys) do not get access \footnote{Centralized entities (like Uber) decide the accessibility of state and action spaces. Typically, to avoid giving away key statistics to competitors, they do not reveal specific information of demand and supply to drivers and choose to reveal aggregate local statistics in the region of taxi driver.} to the joint state $\textbf{s}$, or joint action $\textbf{a}$. Q-function expression for an individual agent $i$ that can observe number of other agents in the zone of agent $i$, i.e., $d_{z_i}(\textbf{s})$ in stochastic games setting will then be:
{\small{\begin{align}
\mathcal{Q}_i(z_i, d_{z_i}(\textbf{s}), a_i) &= {\mathcal R}_i(z_i, d_{z_i}(\textbf{s}),a_i) + \nonumber\\
& \gamma \sum_{z'_i, d'_{z'_i}(\textbf{s}')}\Big[p\big(z'_i,d'_{z'_i}(\textbf{s}') | z_i, {d}_{z_i}(\textbf{s}), a_i\big)  \nonumber \\
&\cdot max_{a'_i} \mathcal{Q}_i(z'_i, d^{'}_{z'_i}(\textbf{s}'),{a}'_i) \Big] \label{joint1}
\end{align}}}
\noindent The above expression is obtained by  considering $(z_i, d_{z_i}(\textbf{s}))$ as the state of agent $i$ in Equation~\ref{QSG}. 

The probability term in Equation~\ref{joint1} is a joint prediction of next zone and number of agents in next zone for agent $i$. Assuming a Naive Bayes approximation for $ p\big(z'_i,d'_{z'_i}(\textbf{s}') | z_i, {d}_{z_i}(\textbf{s}), a_i\big)  $, we have:
{\small \begin{align} 
p\big(z'_i,d'_{z'_i}(\textbf{s}') | z_i, {d}_{z_i}(\textbf{s}), a_i\big)  &\approx p\big(z'_i | z_i, d_{z_i}(\textbf{s}), a_i\big) \cdot \nonumber\\
& \cdot p\big(d'_{z'_i}(\textbf{s}') | z_i, d_{z_i}(\textbf{s}), a_i\big)
\end{align}}
While  the term $p\big(z'_i | z_i, d_{z_i}(\textbf{s}), a_i\big)$ is stationary, the term \\
\noindent $p\big(d'_{z'_i}(\textbf{s}') | z_i, d_{z_i}(\textbf{s}), a_i\big)$ is non-stationary.  $d'_{z'_i}(\textbf{s}')$ is dependent not only on action of agent $i$ but also on actions of other agents (as shown in Figure~\ref{dbn}) and hence $p\big(d'_{z'_i}(\textbf{s}') | z_i, d_{z_i}(\textbf{s}), a_i\big)$ is non-stationary. 

Directly adapting the Q-learning expression of Equation~\ref{eqn:1} to the settings of interest, we have:
{\small \begin{align}
\mathcal{Q}_i(z_i,& d_{z_i}(\textbf{s}), a_i) \leftarrow  \mathcal{Q}_i(z_i, d_{z_i}(\textbf{s}), a_i) + \alpha [r + \nonumber\\
&\gamma \cdot \max_{a'_i} \mathcal{Q}_{i}(z'_i, d_{z'_i}(\textbf{s}'), a'_i) - \mathcal{Q}_i(z_i, d_{z_i}(\textbf{s}), a_i) ]
\end{align}}
Since a part of the transition dynamics are non-stationary, this Q value update results in significantly inferior performance (as shown in experimental results) for existing approaches (Q-learning, DQN, A2C).  This is primarily because prediction of $d'_{z'_i}$ can become biased due to not representing actions of other agents. 
We  account for such non-stationarity by ensuring that prediction of $d'_{z'_i}(\textbf{s}')$ does not get biased and all options for number of agents in a state (that are viable given past data) are feasible. 

The {\em principle of maximum entropy} \cite{jaynes1957information} is employed in problems where we have some piece(s) of information about a probability distribution but not enough to characterize it fully. In fact, this is the  case with the underlying probability distribution of $p\big(d'_{z'_i}(\textbf{s}') | z_i, d_{z_i}(\textbf{s}), a_i\big)$ for all $z'_i$ or in other words normalized $\textbf{d}'(\textbf{s}')$.  { For purposes of easy explainability, we abuse the notation and henceforth refer to the normalized agent population distribution as $\textbf{d}'$ ( i.e., $\sum_{z'_i} {d}'_{z'_i} = 1$) corresponding to actual agent population distribution, $\textbf{d}'(\textbf{s}')$. }

The {\em principle of maximum entropy} states that the best model of a probability distribution is one that assigns an unbiased non-zero probability to every event that is not ruled out by the given data. Intuitively, anything that cannot be explained is assigned as much uncertainty as possible.  Concretely, when predicting a probability distribution, principle of maximum entropy requires that entropy associated with the distribution be maximized subject to the normalization (sum of all probabilities is 1) and known expected value constraints observed in data (for example average density $\bar{d}'_{z'}$ observed in $K_{z'}$ experiences). In case of $\textbf{d}'$, this corresponds to:
{\small \begin{align}
\max  &-\sum_{z'} d'_{z'} log\big(d'_{z'}\big)  &:: [Maximize Entropy]\nonumber\\
\textbf{s.t.} & \sum_{z'} d'_{z'} = 1  &:: [Normalization]\nonumber \\
& \frac{1}{K_{z'}} \sum_{k \in K_{z'}} d'_{z'}(k)  = \bar{d}'_{z'}  &:: [Expected Value] \label{maxent}
\end{align}}

The key challenge is in performing this constrained maximization of entropy for normalized agent density distribution, $\textbf{d}'$ along with a reinforcement learning method.  We achieve this by making two major changes to existing RL methods:

\begin{quote}[\emph{MaximizeEntropy}]: Including a term for entropy of ${\textbf{d}'}$ referred to as $\mathcal{H}_{{\textbf{d}'}}$ as part of the reward term. This will ensure entropy is maximized along with the expected value. \end{quote}

\begin{quote} [\emph{Normalization}]: Normalized prediction is achieved through softmax computation on $p\big(d'_{z'_i}(\textbf{s}') | z_i, d_{z_i}(\textbf{s}), a_i\big)$ prediction. \end{quote}

\begin{quote} [\emph{Expected Value}]: We predict normalized $\textbf{d}'$ given observed  experiences of $d'_{z'_i}(\textbf{s}')$. The objective of the prediction is to minimize mean square loss between predicted and observed value of $d'_{z'_i}$.  Due to minimizing mean square loss, we approximately satisfy the expected value constraint. \end{quote}

\noindent In domains of interest, there are two other key advantages to constrained entropy maximization of $\textbf{d}'$:
\begin{itemize} 
\item \textit{Controlling the non-stationarity}: The domains mentioned in Section~\ref{motivating} have reward/expected reward values that decrease with increasing number of agents (e.g., chances of taxis getting assigned to customers and hence earning revenue are higher when there are fewer taxis in the same zone) . Due to this property, it is beneficial for agents to spread out across zones with demand rather than assemble only in zones with high demand. In other words,  agents taking actions that will maximize the entropy of $\textbf{d}'$ introduces predictability in agent actions and therefore reduces non-stationarity. 
\item \textit{Reduced variance in learned policies}: In MSMPs, there is homogeneity and consistency in learning experiences of agents because the rewards (revenue model) and transitions (allocation of demand to individuals) are determined consistently by a centralized entity (e.g., Uber, Deliveroo ). The only non-stationarity experienced by an agent is due to other agent policies and/or potentially demand, so the impact of better predicting $\textbf{d}'$ and controlling non-stationarity through maximizing entropy minimizes variance experienced by different learning agents.
\end{itemize}

Given their ability to handle non-stationarity better than tabular Q-learning, we implement this idea in the context of DQN~\cite{mnih2015human} and A2C~\cite{mnih2016asynchronous} methods. It should be noted that an experience $e$ in AyMARL is more extensive than in normal RL and is given by $\Big(z_i, d_{z_i}, a_i, r_i, z'_i, d'_{z'_i}\Big)$.

\subsection{Density Entropy based Deep Q-Networks, DE-DQN}
\begin{figure} 
    \centering
    \subfloat[DQN and DE-DQN]{\includegraphics[scale=0.3]{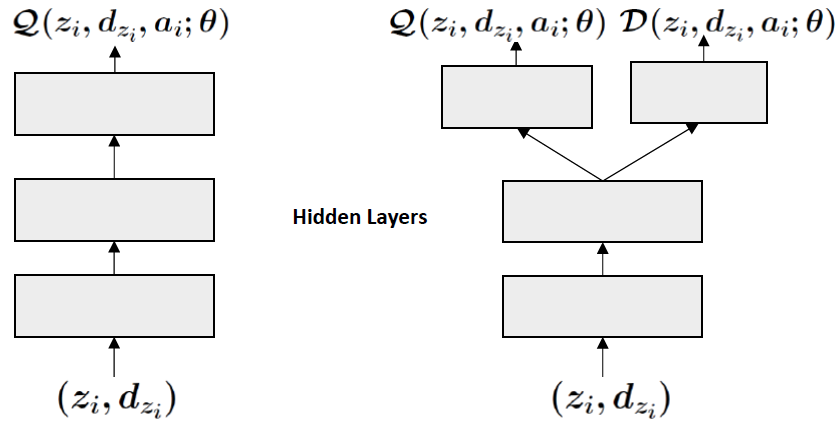}\label{dedqn-net}} \\
     \subfloat[A2C and DE-A2C]{\includegraphics[scale=0.3]{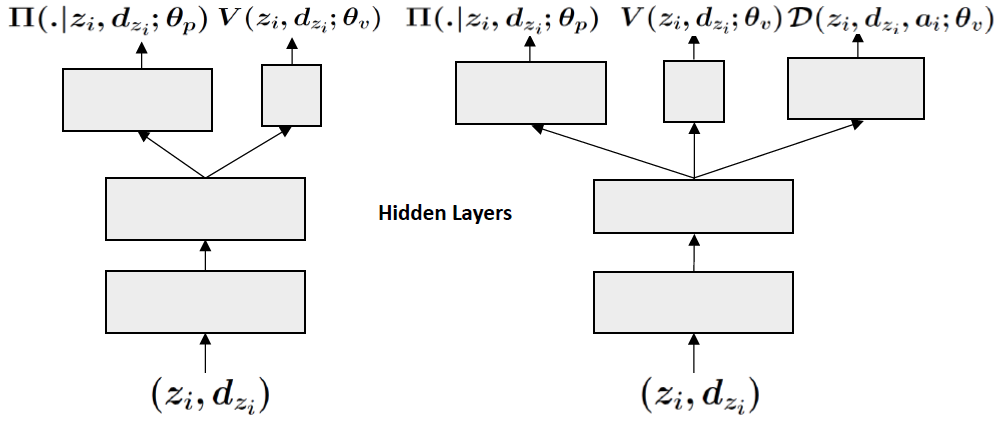}\label{dea2c-net}}
   \caption{DE-DQN and DE-A2C networks}
\end{figure}

We now operationalize the general idea of applying principle of maximum entropy in the context of Deep Q-Networks by modifying the architecture of the neural network and also the loss functions.  We refer to entropy for the predicted future agent population distribution as density entropy. We use $\pmb{\mathpzc{d}}$ to denote the predicted density distribution while $\textbf{d}$ is the true density distribution.

As indicated in Equation~\ref{maxent}, there are three key considerations while applying principle of maximum entropy: (a) maximizing entropy alongside expected reward; (b) ensuring expected value of the computed distribution follows the expected observed samples of $\textbf{d}'$; and (c) finally ensure prediction of $\textbf{d}'$ is normalized. We achieve these considerations using three key steps:
\begin{itemize}
\item \textit{Including entropy in the Q-loss}: This step enables us to maximize entropy alongside expected reward. The Q-loss, $\mathcal{L}_{\theta}^{\mathcal{Q}}$ is updated to include entropy of the predicted agent density for the next step, i.e., $\pmb{\mathpzc{d}}'$. Extending from the description of DQN in Section~\ref{bg:dqn}, $\mathcal{L}_{\theta}^{\mathcal{Q}}$ is given by:

{\small \begin{align}
\resizebox{.86\hsize}{!} {$\mathcal{L}_{\theta}^{\mathcal{Q}} = \mathbb{E}_{(e \sim U(\mathcal{J}))} \Big[\Big(y^{DE-DQN} - \mathcal{Q}\big(z_i, d_{z_i},a_i;\theta\big)\Big)^2\Big]$} \nonumber
\end{align}}

To maximize entropy of $\pmb{\mathpzc{d}}'$ alongside the expected reward, the target value is: 
{\small
\begin{align}
\resizebox{.9\hsize}{!} {$y^{DE-DQN} = r + \beta \mathcal{H}_{\pmb{\mathpzc{d}}'} + \gamma \max_{a'_i} \mathcal{Q}(z'_i, d_{z'_i}, a'_i; \theta^-)$}\nonumber
\end{align}}

Here $\mathcal{H}$ is the density entropy and $\beta$ is a hyperparameter which controls the strength of the entropy.  
\item \textit{Softmax on output layer}: We modify the architecture of the neural network as shown in Figure~\ref{dedqn-net}. This involves introducing a new set of outputs corresponding to prediction of $\textbf{d}'(\textbf{s}')$. By computing softmax on the output, we ensure $\pmb{\mathpzc{d}}'$ is normalized.
\item \textit{Introduce a new density loss, $\mathcal{L}_{\theta}^{\mathcal{D}}$}: This step enables us to minimize loss in prediction of agent density for the next step. Specifically, by minimizing loss we ensure that density entropy maximization occurs subject to observed samples of $\textbf{d}'$.  The independent learner (IL) agent gets to observe only the local density, $d_{z'_i}(\textbf{s}')$ and not the full density distribution $\textbf{d}'$. Hence, we compute mean squared error (MSE) loss, $\mathcal{L^D_{\theta}}$ for the local observation.

{\small
\begin{align}
\mathcal{L_{\theta}^D} = \mathbb{E}_{(e \sim U(\mathcal{J}))} [(d_{z'_i} -\mathcal{D}(z'_i|z_i, d_{z_i}, a_i;\theta))^2] \label{d-loss}
\end{align}}

$\mathcal{D}(z_i,d_{z_i}, a_i; \theta)$ is the density prediction vector and $\mathcal{D}(z'_i|z_i,d_{z_i}, a_i; \theta)$ is the predicted density in zone $z'_i$ if action $a_i$ is taken in state $(z_i, d_{z_i})$. DE-DQN optimizes a single combined loss with respect to the joint parameter $\theta$, $\mathcal{L}_\theta = \mathcal{L}_{\theta}^{\mathcal{Q}} + \lambda \mathcal{L}_{\theta}^{\mathcal{D}}$. $\lambda$ is the weightage term used for the density prediction loss. 
\end{itemize}

\subsection{Density Entropy based A2C, DE-A2C}
Figure \ref{dea2c-net} depicts how A2C network can be modified to DE-A2C network. Similar to DE-DQN, DE-A2C also considers density entropy in value function and  a density loss. In addition, DE-A2C has to consider policy network loss and this is the main difference between DE-A2C and DE-DQN. 

DE-A2C maintains a policy network $\pi(.|z_i, d_{z_i};\theta_p)$ parameterized by $\theta_p$ and a value network parameterized  by $\theta_v$. Value network maintains a value function output $V(z_i,d_{z_i};\theta_v)$ and a density prediction output $\mathcal{D}(z_i,d_{z_i}, a_i; \theta_v)$.  $R$ is $k$-step return from the experience. While computing the value function loss $\mathcal{L}_{\theta_v}^V$, density entropy is included as follows

{\small \begin{align*}
\mathcal{L}_{\theta_v}^V = \mathbb{E}_{(e \sim U(\mathcal{J})} [(R +  \beta \mathcal{H}_{\pmb{\mathpzc{d}}'} - V(z_i,d_{z_i};\theta_v))^2] \label{v-loss-a2cde}\
\end{align*} }
Density prediction loss $\mathcal{L}_{\theta_v}^{\mathcal{D}}$ can be computed as given in equations \ref{d-loss}. The value network loss $\mathcal{L}_{\theta_v}$ is the combination of value loss and density loss, $\mathcal{L}_{\theta_v} = \mathcal{L}_{\theta_v}^V + \lambda \mathcal{L}_{\theta_v}^{\mathcal{D}}$. Similarly, density entropy is included in the policy network loss as follows.
{\small \begin{align*}
\mathcal{L}_{\theta_p} = \mathbb{E}_{(e \sim U(\mathcal{J})} [\nabla_{\theta_p} \text{log} \pi (a_i|z_i, d_{z_i};\theta_p) \\ (R + \beta \mathcal{H}_{\pmb{\mathpzc{d}}'} - V(z_i,d_{z_i};\theta_v))]
\end{align*}}

\section{Experiments}
In this section, we demonstrate that our approaches that employ density entropy alongside Q-function or value function are able to outperform leading independent RL approaches (DQN, A2C, Q-Learning) on Multi-agent Sequential Matching Problems (MSMPs).  DQN, A2C and Q-Learning serve as lower bound baselines on performance. 
We also compare our results with Mean Field Q (MF-Q) learning algorithm \cite{pmlrv80yang18d} which is a centralized learning decentralized execution algorithm. MF-Q computes target values by using previous iteration's mean action (mean of actions taken by neighboring agents), but in our experimental domain the previous mean actions are different for different zones, hence we use the modified MF-Q algorithm (MMF-Q) where agents present in same zone are considered neighbors and previous iteration's mean actions of every zone is available to the all the learning agents. As MMF-Q agents have more information than our DE algorithms, it serves as the upper bound baseline on performance. 

We first provide results on a taxi simulator that is validated on a real world taxi dataset. Second, we provide results on a synthetic online to offline matching simulator (that is representative of problems in Section~\ref{motivating}) under various conditions.

\begin{figure}[h]
    \centering
\includegraphics[scale=0.2]{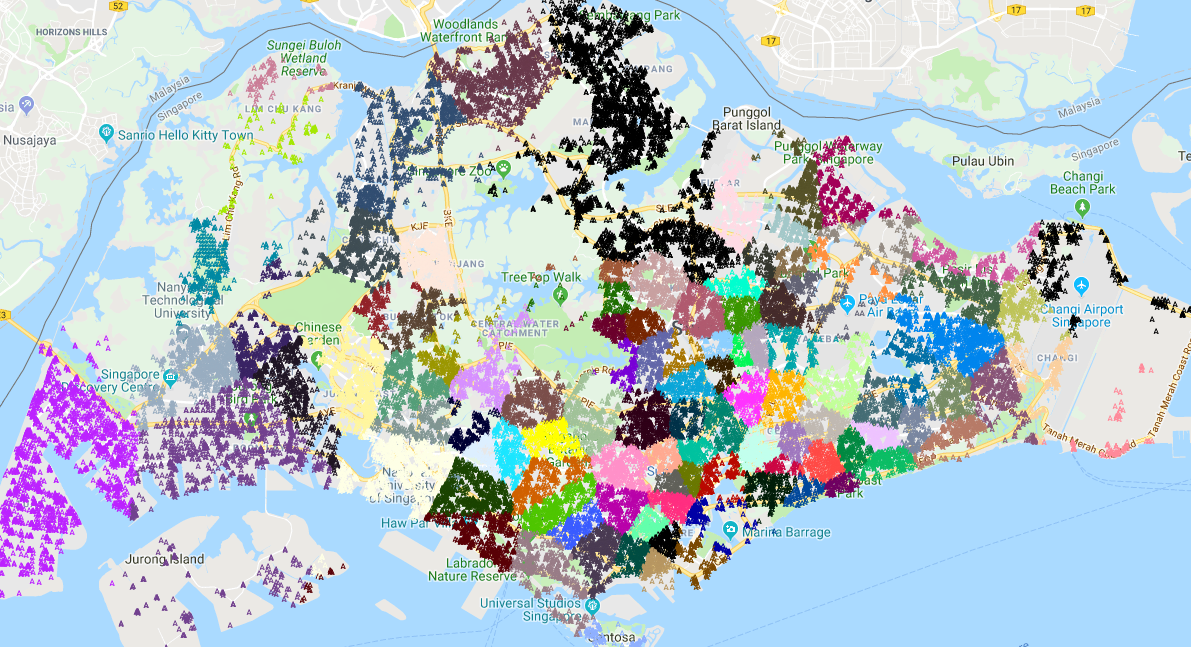}
   \caption{Road network of Singapore divided into zones}
\label{sgp-zones}
\end{figure}
\noindent \textbf{Taxi Simulator:}
Figure \ref{sgp-zones} shows the map of Singapore where road network is divided into zones.  First, based on the GPS data of a large taxi-fleet, we divided the map of Singapore into 111 zones \cite{verma2017rltaxi}. Then we used the real world data to compute the demand between two zones and the time taken to travel between the zones. Since we focus on independent learners, our learning is very scalable. However, simulating thousands of learning agents at the same time requires extensive computer resources and with the academic resources available, we could not perform a simulation with thousands of agents. Hence, we computed proportional demand for 100 agents and simulated for 100 agents.   

\noindent \textbf{Synthetic Online to Offline Service Simulator: } We generated synthetic data to simulate various combinations of demand and supply scenarios in online to offline service settings described in Section~\ref{motivating}. We used grid world and each grid is treated as a zone. Demands are generated with a \textit{time-to-live} value and the demand expires if it is not served within \textit{time-to-live} time periods. Furthermore, to emulate the real-world jobs, the revenues are generated based on distance of the trip (distance between the origin and destination grids). There are multiple agents in the domain and they learn to select next zones to move to such that their long term payoff is maximized. At every time step, the simulator assigns a trip to the agents based on the agent population density at the zone and the customer demand. In our experiments, we make a realistic assumption that agents do not take action at every time step and they can go to any zone from any zone and time taken to travel between zones is proportional to the distance between the grids.
The revenue of an agent can be affected by features of the domain such as 
\begin{itemize}
\item{Demand-to-Agent Ratio (DAR): The average number of customers per time step per agent.}
\item{Trip pattern: The average length of trips can be uniform for all the zones or there can be few zones which get longer trips (for ex. airports which are usually outside the city) whereas few zones get relatively shorter trips (city center).}
\item{Demand arrival rate: The arrival rate of demand can be either static w.r.t. the time or it can vary with time (dynamic arrival rate).}
\end{itemize}
We performed exhaustive experiments on the synthetic dataset where we simulated different combinations of these features.

\subsection{Implementation Details}
For DQN algorithms we used Adam optimizer \cite{kingma2014adam} with a learning rate of 1e-4, whereas for A2C algorithms we used RMSprop optimizer \cite{tieleman2012lecture} with learning rate of 1e-5 for policy network and 1e-4 for value network. Two hidden layers were used with 256 nodes per layer. Both $\beta$ and $\lambda$ were set to 1e-2. To prevent the network from overfitting, we also used dropout layer with 50\% dropout between hidden layers. For DQN algorithms, we performed $\epsilon-$greedy exploration and $\epsilon$ was decayed exponentially. Training is stopped once $\epsilon$ decays to 0.05.

\subsubsection{Computing Density Entropy from Local Observation}
The independent learner agent gets to observe only the local density, $d_{z'_i}$ and not the full density distribution $\textbf{d}'$. This makes computing a cross-entropy loss (or any other relevant loss) for density prediction difficult. Hence, as shown in Equation \ref{d-loss}, we compute MSE loss and to get normalized density prediction $\pmb{\mathpzc{d}}'$, we apply softmax to the density output $\mathcal{D}(z_i, d_{z_i}, a_i;\theta)$. Density entropy can be computed as $
\mathcal{H}_{\pmb{\mathpzc{d}}'} = - \pmb{\mathpzc{d}}' \cdot \text{log}(\pmb{\mathpzc{d}}')
$
\begin{figure}
    \centering
     \includegraphics[scale=0.22]{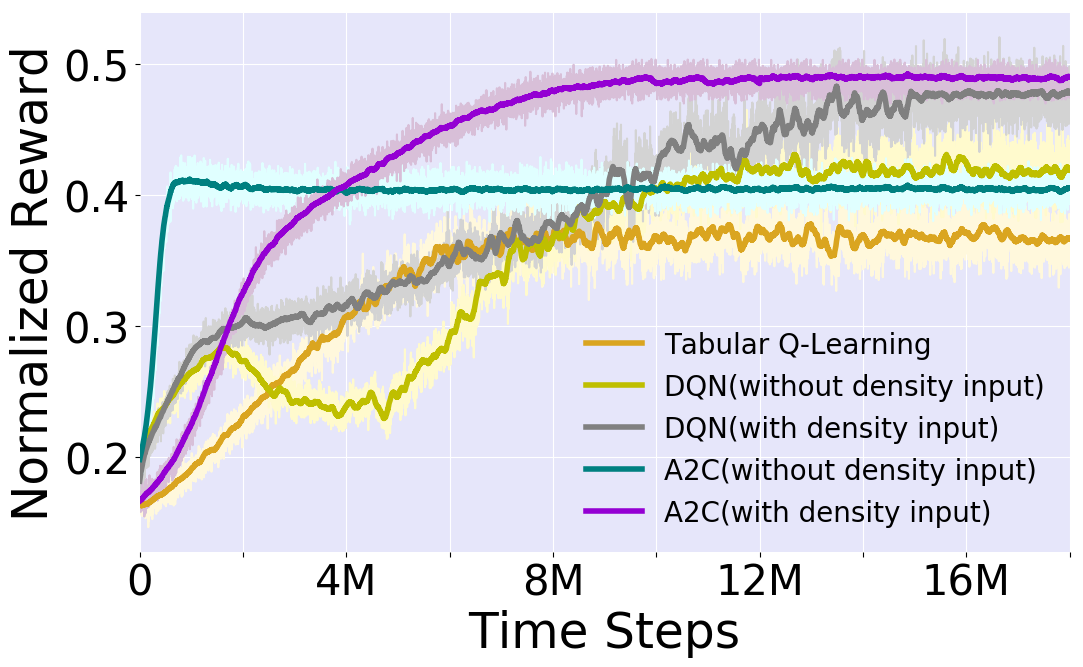}
\caption{Comparison of tabular Q-learning, standard DQN and standard A2C .}  
 \label{vanilla}
\end{figure}

\begin{figure}
    \centering
     \includegraphics[scale=0.22]{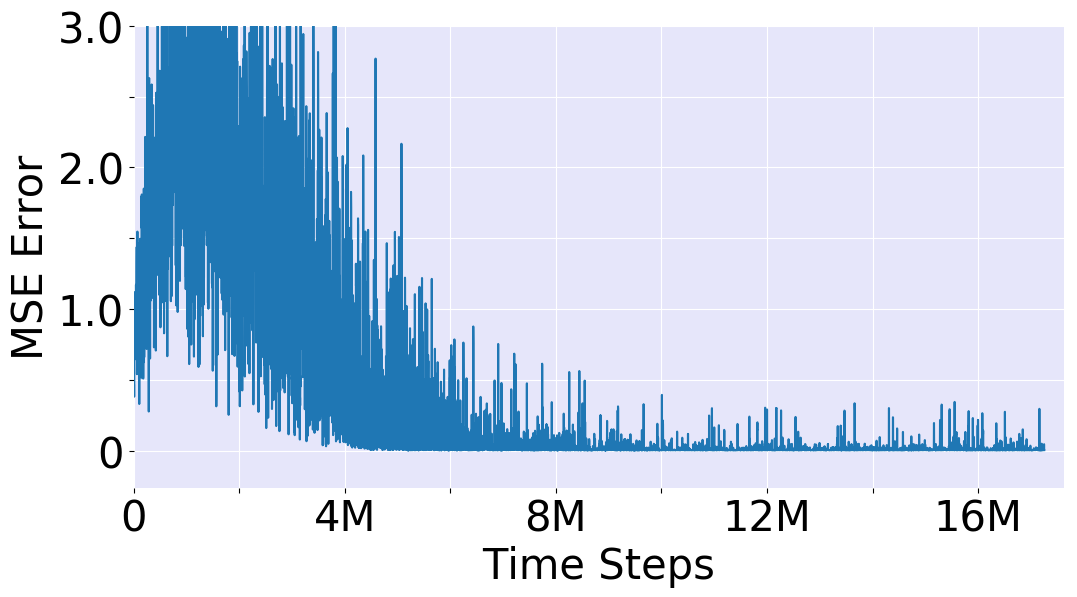}
\caption{Error is density prediction.}  
 \label{d-error}
\end{figure}

\begin{figure*}[t]
   \centering
   \subfloat[Real world dataset]{\label{real-world}\includegraphics[scale=0.159]{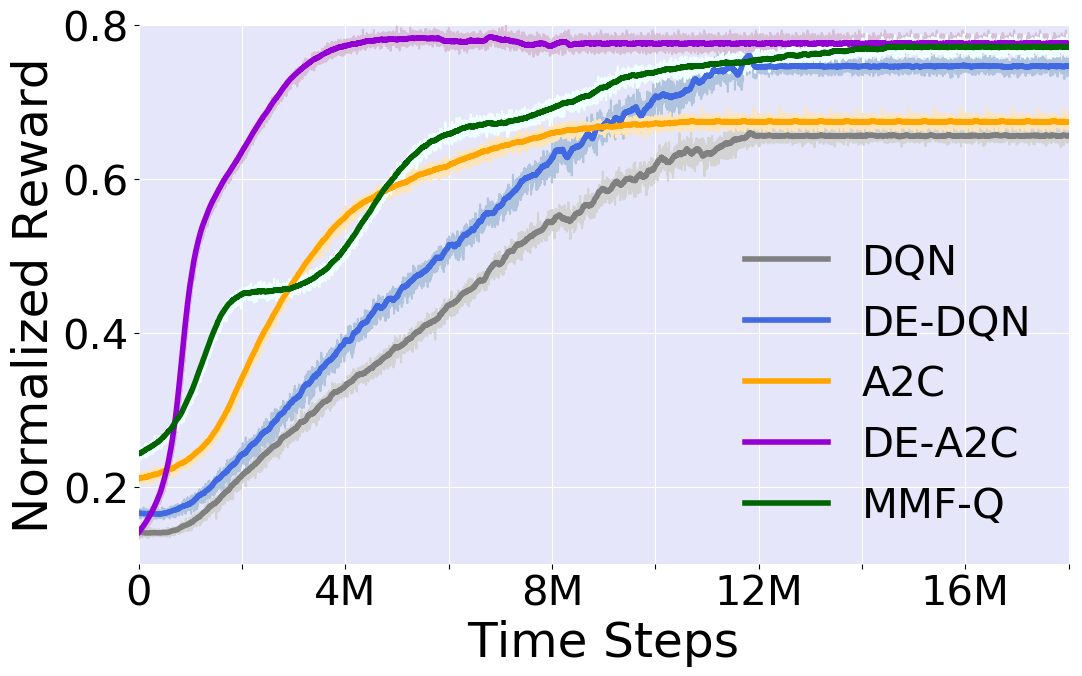} 
       \includegraphics[scale=0.159]{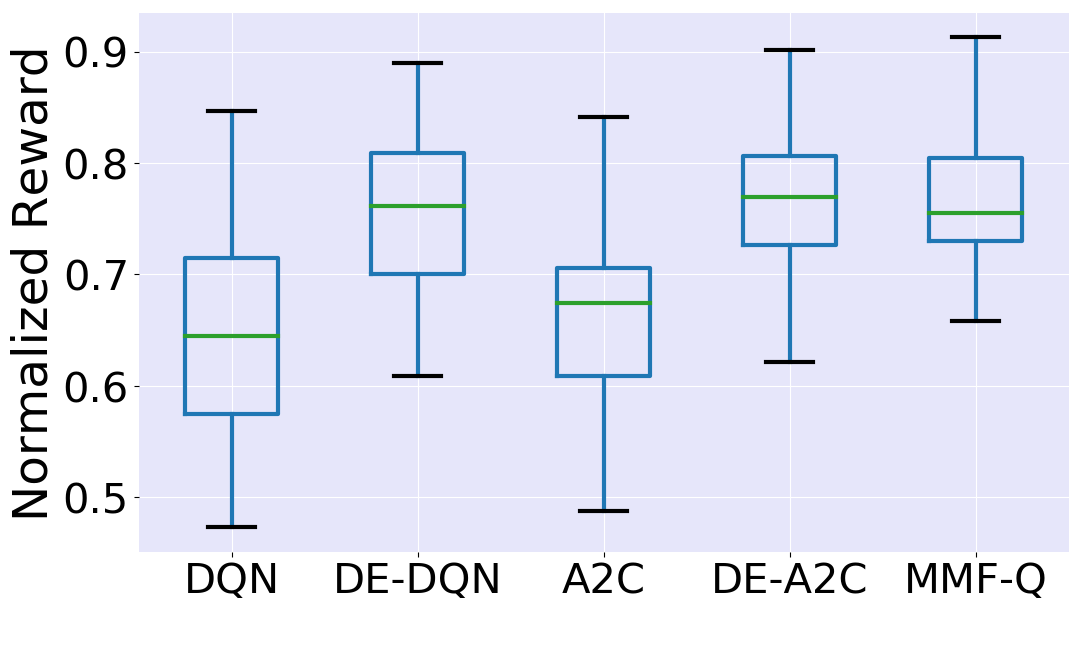}} 
        \subfloat[$\text{DAR}=0.25$ with uniform trip pattern.]{\label{dar25}\includegraphics[scale=0.159]{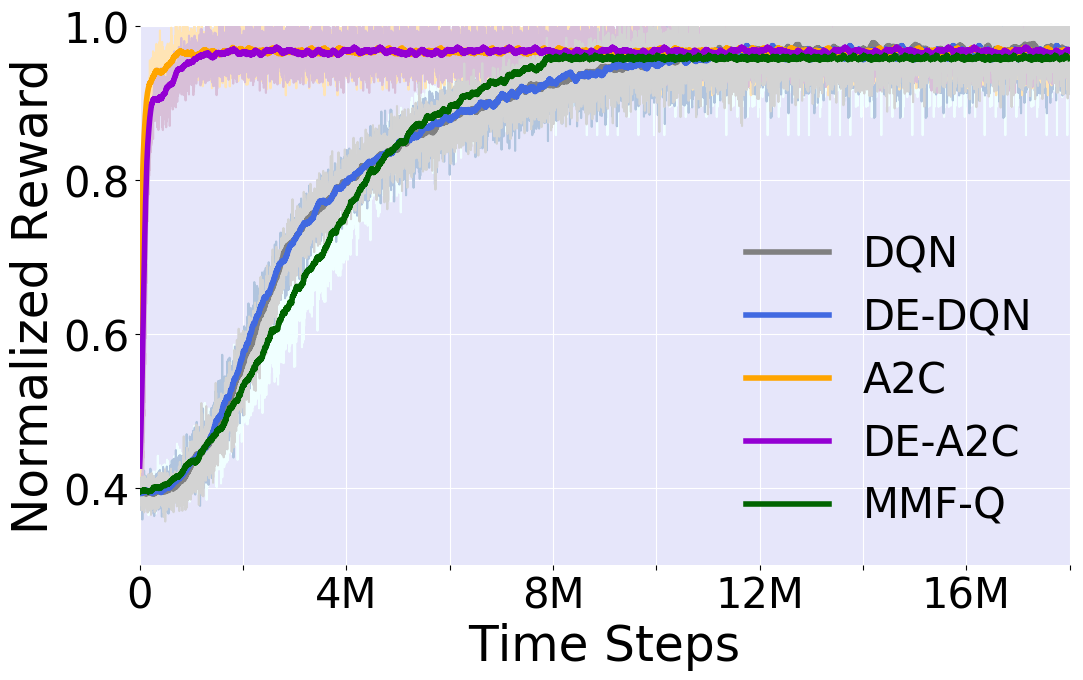} 
       \includegraphics[scale=0.159]{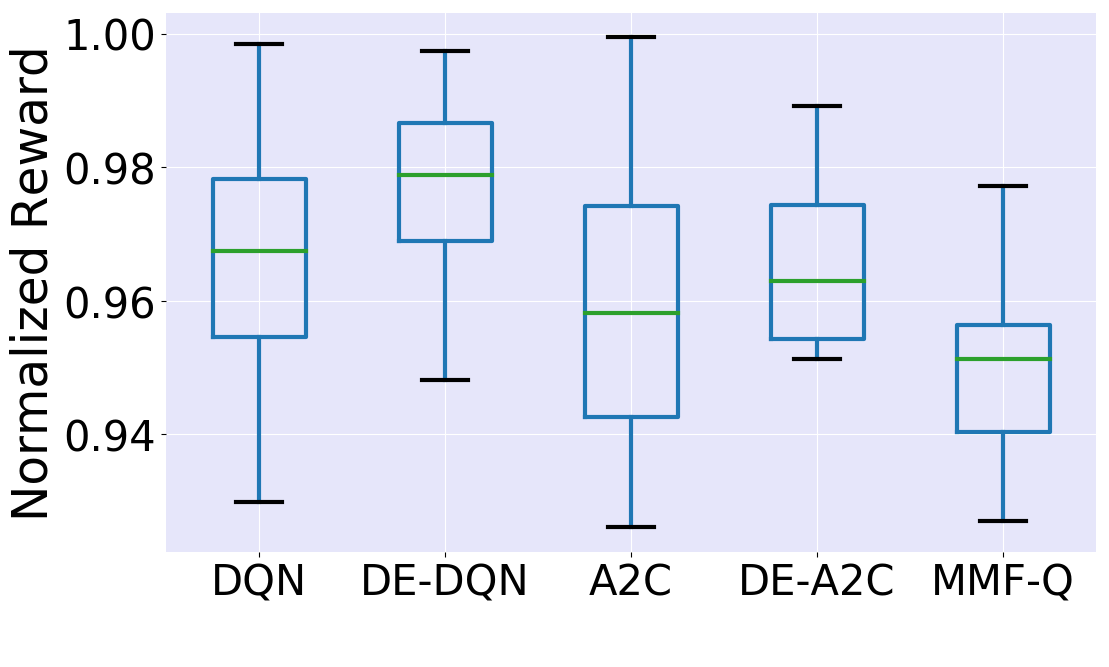}} \\
        \subfloat[$\text{DAR}=0.4$ with dynamic demand arrival rate.]{\label{dar40d}\includegraphics[scale=0.159]{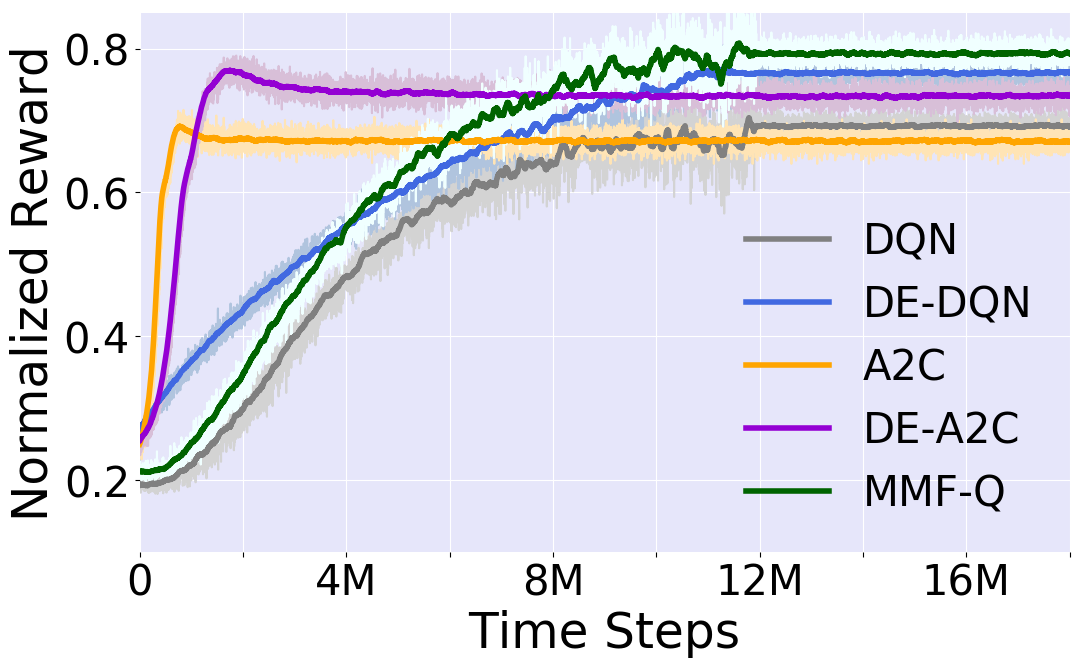} 
       \includegraphics[scale=0.159]{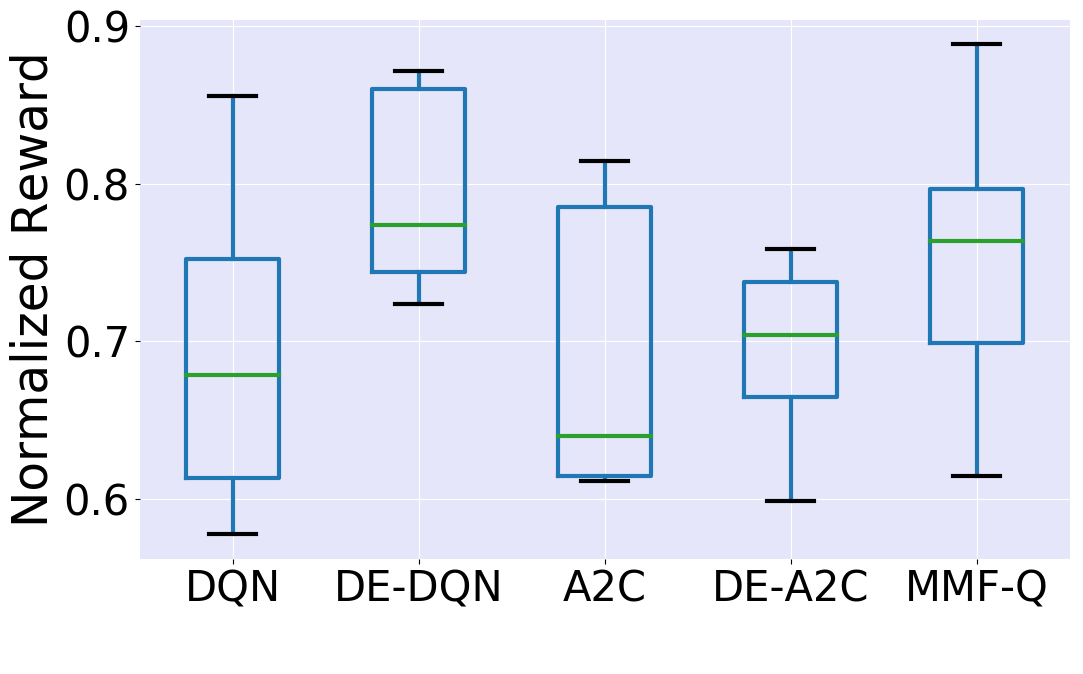}}
\subfloat[$\text{DAR}=0.5$ with non-uniform trips pattern.]{\label{dar50}\includegraphics[scale=0.159]{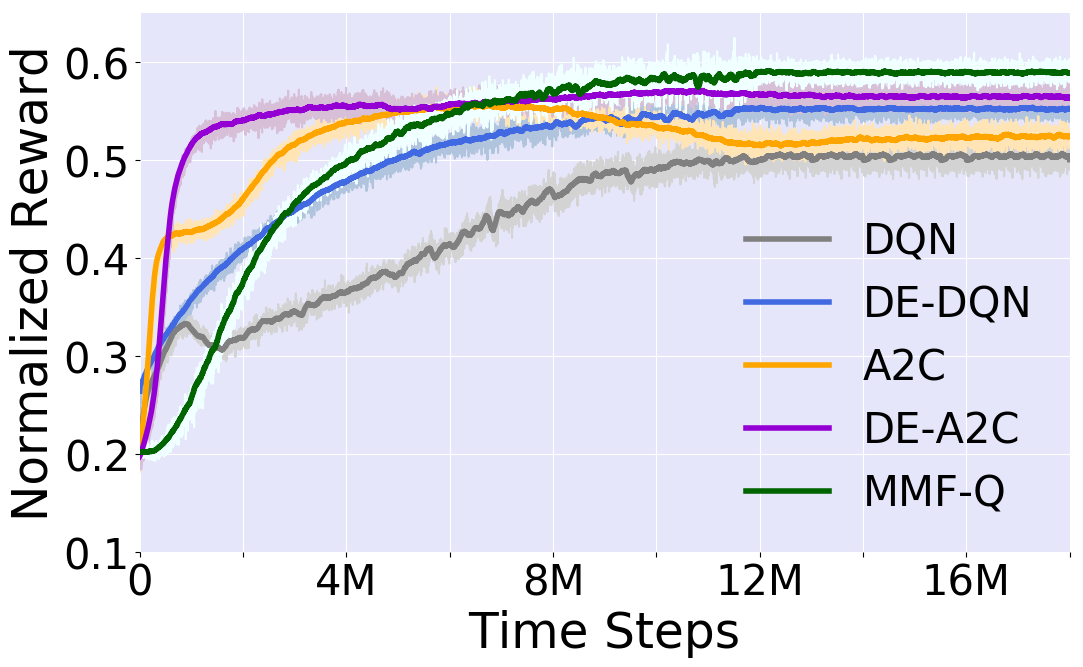}
       \includegraphics[scale=0.159]{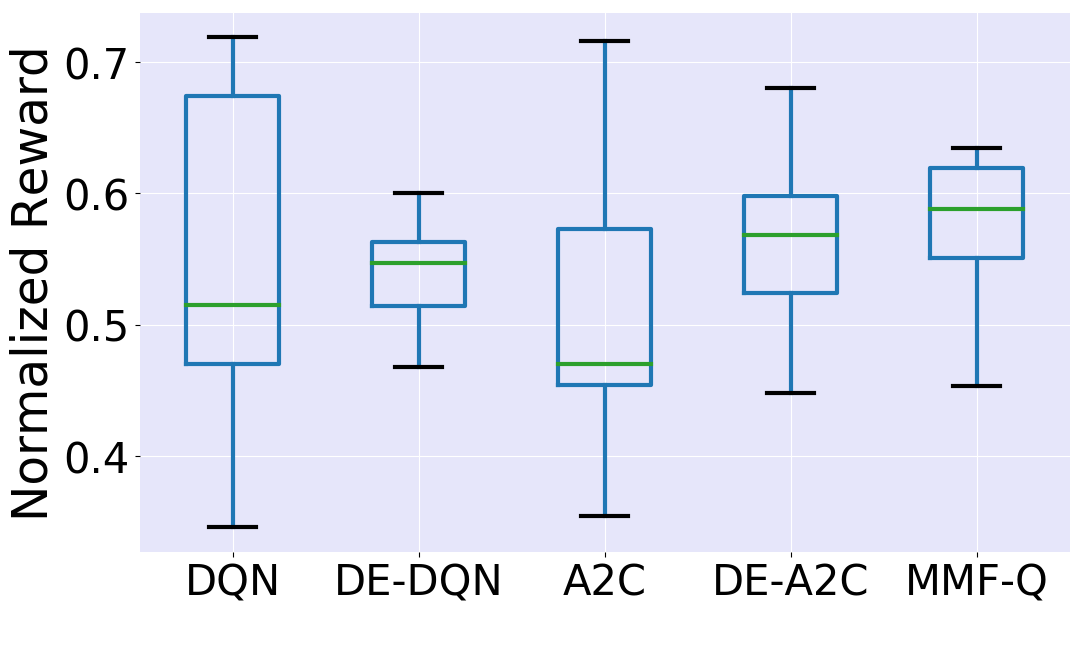}} \\
        \subfloat[$\text{DAR}=0.6$ with uniform trip pattern.]{\label{dar60}\includegraphics[scale=0.159]{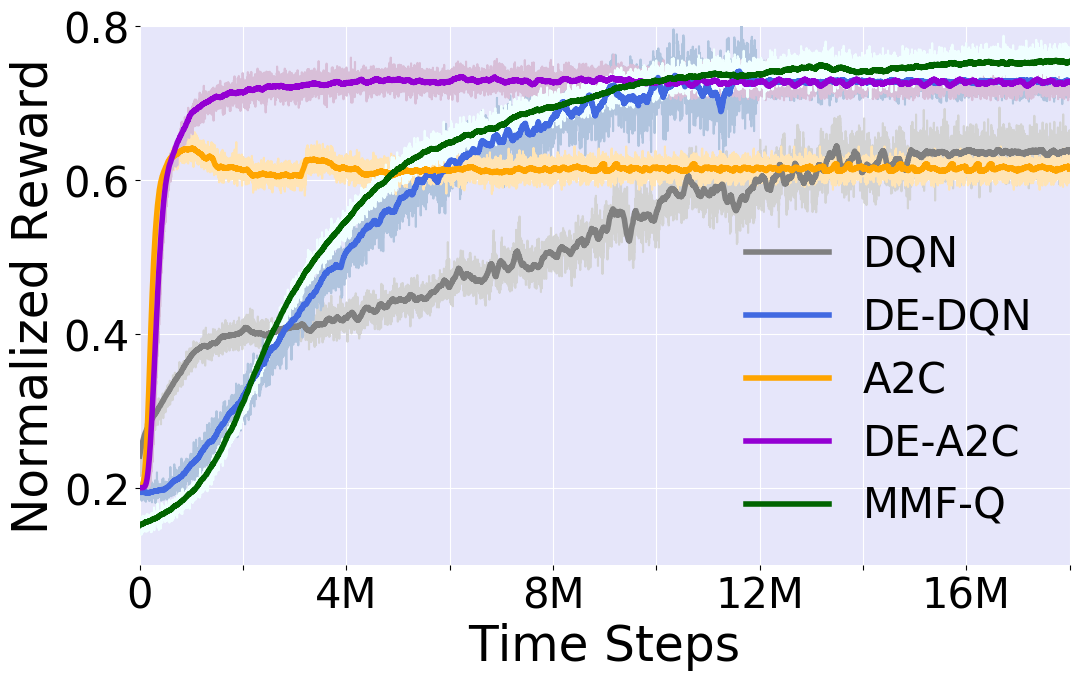} 
       \includegraphics[scale=0.159]{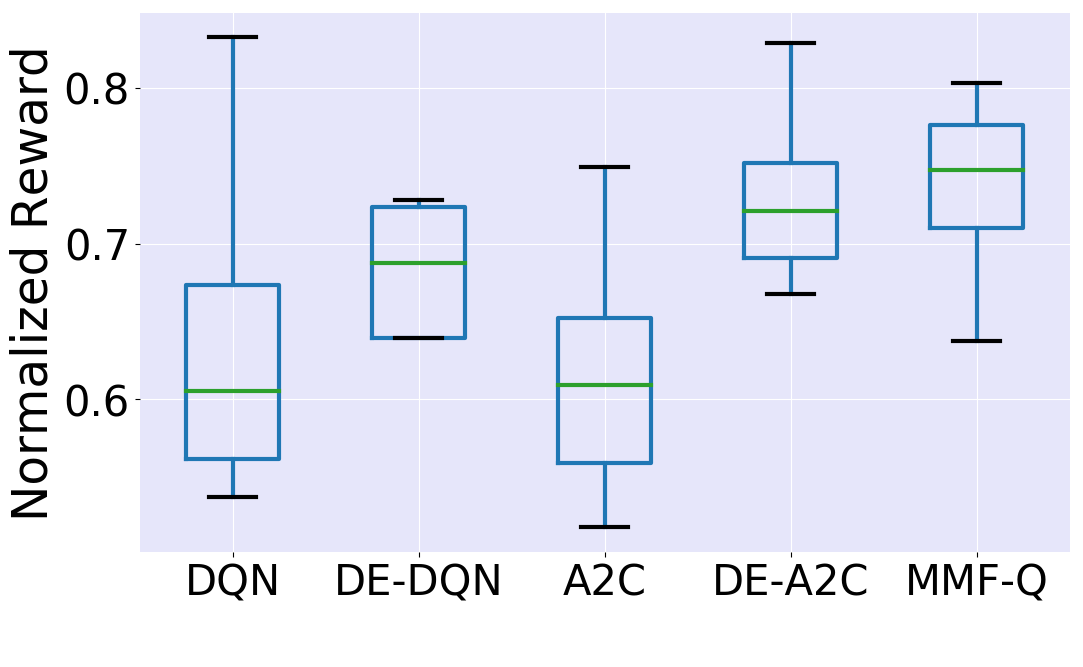}} 
   \subfloat[$\text{DAR}=0.75$ with non-uniform trip pattern.]{\label{dar75}\includegraphics[scale=0.159]{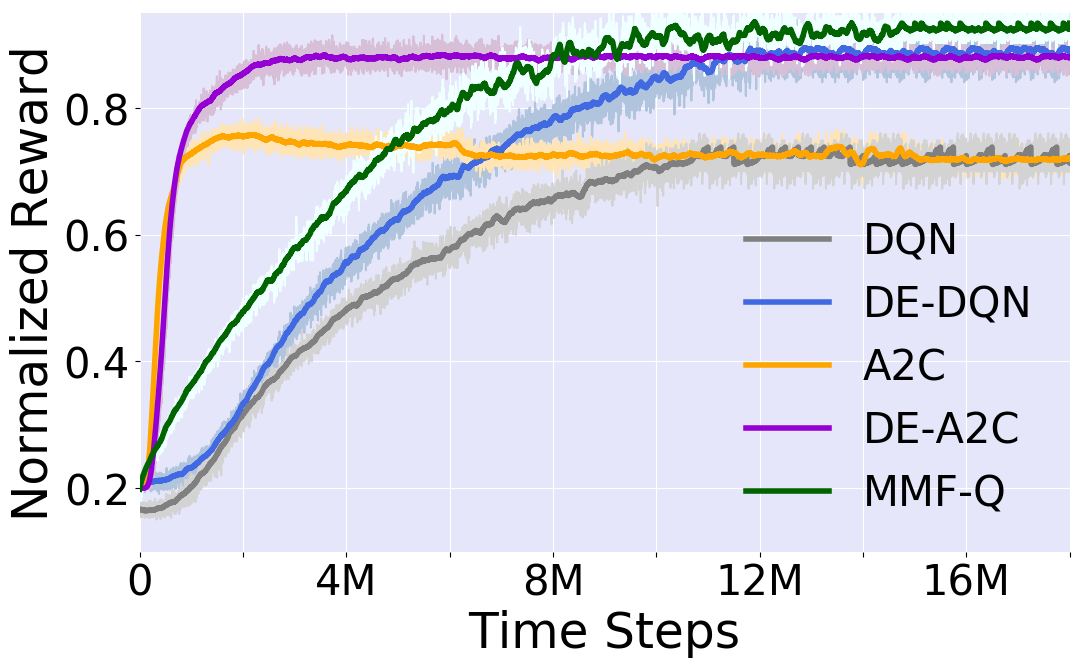} 
       \includegraphics[scale=0.159]{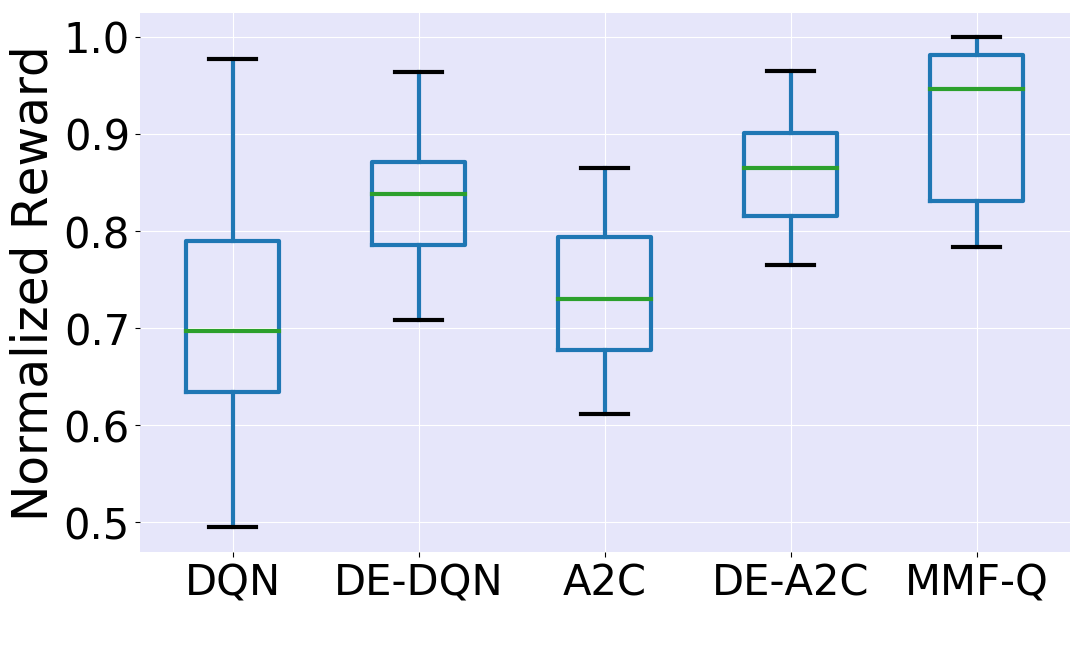}} 
   \caption{Average reward and individual rewards for different experimental setup.} \label{plots}
\end{figure*}

\subsection{Results}
We now benchmark the performance of our learning approaches (DE-DQN and DE-A2C) with respect to DQN, A2C and MMF-Q.  One evaluation period consists of 1000 (1e3) time steps\footnote{A time step represents a decision and evaluation point in the simulator.} and revenue of all the agents is reset after every evaluation period. All the graphs plotted in the upcoming sub-sections provide running average of revenue over 100 evaluation periods (1e5 time steps). 

We evaluated the performance of all learning methods on two key metrics: \\
\noindent (1) Average payoff of all the individual ILs. This indicates social welfare. Higher values imply better performance.  \\
\noindent (2) Variation in payoff of individual ILs after the learning has converged. This is to understand if agents can learn well irrespective of their initial set up and experiences. This in some ways represents fairness of the learning approach.  We use box plots to show the variation in individual revenues, so smaller boxes are better.

To provide a lower benchmark, we first compared our approaches with tabular Q-Learning. We also performed experiments with DQN and A2C algorithms when density input is not provided to the neural network. Figure \ref{vanilla} shows the comparison between tabular Q-learning; DQN with and without agent density input and A2C with and without agent density input where we plot average payoff of all the ILs (social welfare). We used non-uniform trip pattern with $\text{DAR}=0.6$ for this experiment.
We can see that DQN with agent density input and A2C with agent density input perform significantly better than the other three algorithms. For other demand/supply experimental setups also we obtained similar results. Hence, for our remaining experiments we used DQN and A2C with agent density input as our lower baseline algorithm.

In Figure \ref{d-error} we empirically show that the MSE loss between predicted density $\pmb{\mathpzc{d}}$ and true density distribution $\textbf{d}$ converges to a low value ($\approx 0.2$).

For synthetic service simulator, we used multiple combinations of agent population and number of zones (20 agents-10 zones, 50 agents-15 zones, 50-agents-25 zones etc.). 
First plots in Figure \ref{plots} provide comparison of average payoff of all the ILs for the respective experimental setups\footnote{The demand arrival rate is static and trip pattern is uniform for the experimental setups until stated otherwise.}. Second plots uses boxplots to show the variation in the individual average revenues after the learning has converged. Here are the key conclusions:
\begin{itemize}
\item Experiment on real-world dataset (Figure \ref{real-world}) show that DE-DQN and DE-A2C outperformed DQN and A2C approaches by $\approx$10\%.
\item  For low DAR (Figure \ref{dar25}) there is no overall improvement in the social welfare which is expected as with too many agents and too less demand, random action of some or the other agent will serve the demand. However, the variance is lower than the baseline algorithms.
\item With increase in DAR, the performance gap starts increasing ($\approx$ 7\% for $\text{DAR}=0.4$, $\approx$10\% for $\text{DAR}=0.6$ and $\approx$15\% for $\text{DAR}=0.75$) 
\item  DE-DQN and DE-A2C were able to perform as well as MMF-Q even with local observations on the real world dataset validated simulator.  Even on the synthetic simulator, the gap was quite small (about 5\% for $\text{DAR}=0.75$). The gap is lesser for lower values of DAR. 
\item The variation in revenues was significantly\footnote{Differences between DE algorithm and their counterpart base algorithm are statistically significant at 5\% level.} lower for DE-DQN and DE-A2C compared to other approaches, thus emphasizing no unfair advantage for any agents. 
\end{itemize}

\section{Conclusion}
Due to the key advantages of predicting agent density distribution (required for accurate computation of Q-value) and controlling non-stationarity (due to agents changing policies), our key idea of maximizing agent density entropy is extremely effective in Anonymous Multi Agent Reinforcement Learning settings. We were able to demonstrate the utility of our new approaches on a simulator validated on real world data set and two other generic simulators both with respect to overall value for all agents and fairness in values obtained by individual agents.

\section{Acknowledgements}
This research is supported by the National Research Foundation Singapore under its Corp Lab @ University scheme and Fujitsu Limited as part of the A*STAR-Fujitsu-SMU Urban Computing and Engineering Centre of Excellence.

\bibliographystyle{aaai}  
\bibliography{marl}  

\end{document}